# Hyperproperty-Constrained Secure Reinforcement Learning

Ernest Bonnah
Bayloy University
Waco, Texas, USA

Luan Viet Nguyen
University of Dayton
Dayton, Ohio, USA

Khaza Anuarul Hoque
University of Missouri-Columbia
Columbia, Missouri, USA

## ABSTRACT

Hyperproperties for Time Window Temporal Logic (HyperTWTL) is a domain-specific formal specification language known for its effectiveness in compactly representing security, opacity, and concurrency properties for robotics applications. This paper focuses on HyperTWTL-constrained secure reinforcement learning (SecRL). Although temporal logic-constrained safe reinforcement learning (SRL) is an evolving research problem with several existing literature, there is a significant research gap in exploring security-aware reinforcement learning (RL) using hyperproperties. Given the dynamics of an agent as a Markov Decision Process (MDP) and opacity/security constraints formalized as HyperTWTL, we propose an approach for learning security-aware optimal policies using dynamic Boltzmann softmax RL while satisfying the Hyper-TWTL constraints. The effectiveness and scalability of our proposed approach are demonstrated using a pick-up and delivery robotic mission case study. We also compare our results with two other baseline RL algorithms, showing that our proposed method outperforms them.

## 1 INTRODUCTION

Reinforcement learning (RL) has succeeded tremendously in many complex decision-making tasks. However, in real-world applications, safety is a major concern, and the area of safe RL (SRL) is still in its early stages [21]. Several model-free RL methods have been proposed in conjunction with temporal logic specifications, such as using Linear Temporal Logic (LTL) [13, 14], Signal Temporal Logic (STL) [26, 36, 38], Metric Interval Temporal Logic (MITL) [28, 40], and Time Window Temporal Logic (TWTL) [5, 6] to *guarantee* safe learning of optimal control policies for autonomous systems that interact with environments subject to stochastic uncertainties [39]. Interestingly, none of these existing works in temporal logic-constrained SRL enforces security/opacity policies (in addition to safety) while learning optimal control policies. With increasing cybersecurity risks in robotics [19, 29, 41, 42], there is a pent-up need to explore this research direction.

Traditional temporal logics, such as LTL, MTL, STL, and TWTL, can only express trace properties, i.e., the specified properties involve reasoning about individual executions or traces. This limits their application to many other domains, which require reasoning about multiple traces. *Hyperproperties* extend traditional trace properties to express properties of sets of traces and thus can directly specify a wide range of important properties such as information-flow security, consistency models in concurrent computing [9, 17], robustness levels in cyber-physical systems [8, 20], opacity, and also service level agreements [16]. Motivated by the expressiveness of hyperproperties, we introduce a hyperproperties-constrained secure (SecRL) in this paper. We use hyperproperties for the time window temporal logic (HyperTWTL) [11, 12] for specifying the security constraints to guide the constrained learning process of the optimal policies. HyperTWTL is known for its *compactness* and *effectiveness* in specifying a wide range of safety and security properties for robotic missions. For instance, consider a non-interference hyperproperty that requires that "*for any pair of traces $\pi_1$ and $\pi_2$ of a system, low-security variables $L$ should always be independent of high-security variables $H$ within the time bound $[0, 10]$.*" *This requirement $\varphi$ can be expressed using HyperMTL formalisms as* $\forall \pi_1 \forall \pi_2 \cdot (\bigvee_{i=0}^{10} \mathbf{G}_{[i,i+10]}(L_{\pi_1} = L_{\pi_2})) \wedge (\bigvee_{i=0}^{10} \mathbf{G}_{[i,i+10]}(H_{\pi_1} = H_{\pi_2}))$, *which requires 29 operators. The same specification can be expressed in HyperTWTL using only 10 operators as* $\forall \pi_1 \forall \pi_2 \cdot [\mathbf{H}^{10} \ L_{\pi_1} = \mathbf{H}^{10} \ L_{\pi_2}]^{[0,10]} \rightarrow [\mathbf{H}^{10} \ H_{\pi_1} = \mathbf{H}^{10} \ H_{\pi_2}]^{[0,10]}$. For the RL algorithm, our method employs the Dynamic Boltzmann softmax approach [33] to maximize the expected rewards sum subject to hyperproperty constraints during learning optimal policies. Dynamic Boltzmann softmax RL is known for having a good convergence guarantee [4]. We model the dynamics of the interaction between the agent and the environment as a Markov Decision Process (MDP) with unknown transition probabilities. Our approach converts an alternation-free HyperTWTL[1] formula into a Deterministic Finite automaton (DFA) and combines the generated automaton with the input MDP to generate a product (and timed) MDP. The resulting product MDP is then used to learn optimal policies using dynamic Boltzmann softmax RL that satisfies the given security constraints formalized as HyperTWTL formulae. To demonstrate the effectiveness and scalability of our approach, we formalize two interesting

---

[1] Our proposed approach in this paper is limited to the alternation-free fragment of HyperTWTL

opacity/security requirements as constraints for a pick-up and delivery mission using HyperTWTL. We compare the results obtained with two other RL algorithms: Q-learning and a modified Dyna-Q algorithm, which shows that our proposed method outperforms them.

## 2 PRELIMINARIES

Let $AP$ be a finite set of *atomic propositions* and $\Sigma = 2^{AP}$ be the *alphabet*, where each member of $\Sigma$ is called an *event*. We define a timed trace $t$ as a finite sequence of events from $\Sigma^*$, i.e., $t = (\tau_i, e_i), (\tau_{i+1}, e_{i+1}), \cdots (\tau_n, e_n) \in (\mathbb{Z}_{\geq 0} \times \Sigma)^*$ where $\tau_i \tau_{i+1} \cdots \tau_n \in \mathbb{Z}_{\geq 0}$ is a sequence of non-negative integers denoting *time-stamps* and the indices $i, n \in \mathbb{Z}_{\geq 0}$ denote *time-points*. We require $\tau_i = 0$, $\tau_i \leq \tau_{i+1}$, and for all $i, 0 \leq i \leq n$. For each timed trace $t$, by $t[i].e$, we mean $e_i$ and by $t[i].\tau$ we mean $\tau_i$. We now define an indexed timed trace as a pair $(t, p)$ where $p \in \mathbb{Z}_{\geq 0}$ is called a pointer. Indexed timed traces allow traversing a given trace by moving the pointer. Given an indexed timed trace $(t, p)$ and $m \in \mathbb{Z}_{\geq 0}$, let $(t, p) + m$ denote the resulting trace $(t, p + m)$.

### 2.1 HyperTWTL

HyperTWTL [11, 12] specifies hyperproperties for TWTL [37] by extending the classical TWTL with quantification over multiple and concurrent execution traces. This section briefly presents the syntax and semantics of HyperTWTL. The syntax and semantics of HyperTWTL are described as follows while assuming that the timestamps of all the quantified traces are synchronous, i.e., all the timestamps of traces match at each point in time. The set of formulae in HyperTWTL is inductively defined by the following syntax:

$$\varphi := \exists \pi \cdot \varphi \mid \forall \pi \cdot \varphi \mid \phi$$

$$\phi := \mathbf{H}^d a_\pi \mid \mathbf{H}^d \neg a_\pi \mid \phi_1 \wedge \phi_2 \mid \neg \phi \mid \phi_1 \odot \phi_2 \mid [\phi]^{[x,y]}$$

where $a$ is an atomic proposition in $AP$ and $\pi$ is a trace variable in the set of trace variables $\mathcal{V}$. The quantified traces $\exists \pi$ and $\forall \pi$ are interpreted as "there exists at least a trace $\pi^*$" and "for all the traces $\pi^*$" respectively. The operators $\mathbf{H}^d, \odot$ and $[\ ]^{[x,y]}$ represent the hold operator with $d \in \mathbb{Z}_{\geq 0}$, concatenation operator and within operator respectively with a discrete-time constant interval $[x, y]$, where $x, y \in \mathbb{Z}_{\geq 0}$ and $y \geq x$, respectively. $\wedge$ and $\neg$ are the conjunction and negation Boolean operators, respectively. The disjunction operator ($\vee$) can be derived from the negation and conjunction operators. Likewise, the implication operator ($\rightarrow$) can also be derived from the negation and disjunction operators.

The satisfaction of a HyperTWTL formula $\varphi$ is a binary relation $\models$, that relates a HyperTWTL formula $\varphi$ with trace set $\mathbb{T}$ over a trace assignment $\Pi$. The semantics of (synchronous) HyperTWTL [11, 12] are presented in Table 1. We define an assignment $\Pi : \mathcal{V} \rightarrow (\mathbb{Z}_{\geq 0} \times \Sigma)^* \times \mathbb{Z}_{\geq 0}$ as a partial function mapping trace variables to indexed timed traces. We therefore denote the mapping of the trace variable to an index timed trace $(t, p)$ as $\Pi[\pi \rightarrow (t, p)]$. Thus, by $\Pi(\pi) = (t, p)$, we mean the event from the timed trace $t$ at the position $p$ is currently used in the analysis of trace $\pi$. Given a set of traces denoted as $\mathbb{T}_{[i,j]}$, we say the evaluation of all the traces in the given set against a formula starts from the time-point $i \geq 0$ up to and includes the time-point $j \geq i$. We use $(\Pi) + k$ as the $k^{th}$ successor of $\Pi$, i.e., the $k^{th}$ timed event of a mapped trace reached after moving $k$ steps across $\Pi$. The hold operator $\mathbf{H}^d a_\pi$ states that the proposition $a$ is to be repeated for $d$ time units in trace $\pi$. Similarly $\mathbf{H}^d \neg a_\pi$, requires that for $d$ time units the proposition $a$ should never hold in trace $\pi$. The trace set $\mathbb{T}$ must satisfy both sub-formulae $\phi_1$ and $\phi_2$ in $\phi = \phi_1 \wedge \phi_2$ while in $\phi = \neg \phi$, $\mathbb{T}$ does not satisfy the given formula. A given formula with a concatenation operator in the form $\phi = \phi_1 \odot \phi_2$ specifies that every $t \in \mathbb{T}$ should satisfy $\phi_1$ first and then immediately $\phi_2$ must also be satisfied with one-time unit difference between the end of execution of $\phi_1$ and the start of execution of $\phi_2$. Given $\phi = [\phi]^{[x,y]}$, the trace set $\mathbb{T}$ must satisfy $\phi$ within the bound $[x, y]$. We now define the current instant denoted as $(\Pi)^{now}$ and the $j^{th}$ instant denoted as $(\Pi)^j$ given $\Pi$ ([10]) as $(\Pi)^{now} = \max_{\pi \in dom(\Pi)} = \{t[p].\tau \mid \text{ for } \Pi(\pi) = (t, p)\}$ and $(\Pi)^j = \min_{\pi \in dom(\Pi)} = \{t[p + j].\tau \mid \text{ for } \Pi(\pi) = (t, p)\}$, respectively.

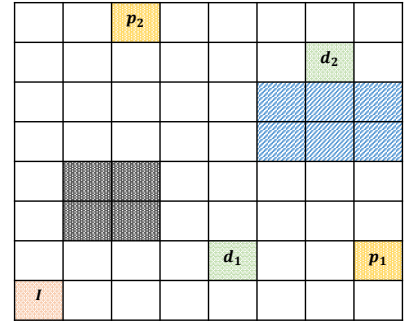

**Figure 1: Environment with initial (beige), pick-up (yellow), delivery (green), obstacles (blue), reward (grey) regions**

***HyperTWTL Execution Deadline.*** The satisfaction of a Hyper-TWTL formula can be verified within bounded time. We denote the maximum time needed to satisfy $\varphi$ by $||\varphi||$, which can be recursively computed as follows:

$$||\varphi|| = \begin{cases} ||\varphi|| & \text{if} \quad \varphi \in \{\exists \pi \cdot \varphi, \forall \pi \cdot \varphi\} \\ d & \text{if} \quad \varphi \in \{\mathbf{H}^d a_\pi, \mathbf{H}^d \neg a_\pi\} \\ max(||\varphi_1||, ||\varphi_2||) & \text{if} \quad \varphi \in \{\varphi_1 \wedge \varphi_2, \varphi_1 \vee \varphi_2\} \\ ||\varphi_1|| & \text{if} \quad \varphi = \neg \varphi_1 \\ ||\varphi_1|| + ||\varphi_2|| + 1 & \text{if} \quad \varphi = \varphi_1 \odot \varphi_2 \\ y & \text{if} \quad \varphi = [\varphi_1]^{[x,y]} \end{cases} \quad (1)$$

### 2.2 Applications of HyperTWTL

To demonstrate the feasibility of HyperTWTL we use a case study that resembles a pick-up and delivery mission. As shown in Figure 1, an $8 \times 8$ environment for the pick-up and delivery mission is made up of an initial state $I$ (beige), two pick-up locations $p_1$ and $p_2$ (yellow), two delivery locations, $d_1$ and $d_2$ (green), reward locations (grey) and obstacle locations, $O$ (blue). On each mission, delivery drones are required to perform pick-up tasks from $p_1$ or $p_2$ within the time limit $[0, T_1]$ and delivery tasks in $d_1$ or $d_2$ within bound $[T_2, T_3]$ while avoiding all obstacles $O$. Given the pick-up and delivery locations, a set of missions consisting of a combination of these locations is defined such that $\varphi_{tasks} = \{\varphi_{p_1 d_1}, \varphi_{p_1 d_2}, \varphi_{p_2 d_1}, \varphi_{p_2 d_2}\}$ where $\varphi_{p_i d_j}$ encodes the mission where pick-up and delivery locations are $p_i$ and $d_j$ given $i, j \in \{1, 2\}$. Based on this pick-up and delivery mission, we consider two hyperproperties, namely, opacity

**Table 1: Semantics of HyperTWTL**

| | | |
|---|---|---|
| $(\mathbb{T}, \Pi) \models \exists \pi. \varphi$ | iff | $\exists t \in \mathbb{T} \cdot (\mathbb{T}, \Pi[\pi \to (t, 0)]) \models \varphi$ |
| $(\mathbb{T}, \Pi) \models \forall \pi. \varphi$ | iff | $\forall t \in \mathbb{T} \cdot (\mathbb{T}, \Pi[\pi \to (t, 0)]) \models \varphi$ |
| $(\mathbb{T}, \Pi) \models \mathbf{H}^d a_\pi$ | iff | $a \in t[p].e$ for $(t, p) = \Pi(\pi), \forall p \in \{i, ..., i+d\} \wedge (t[i+n].\tau - t[i].\tau) \geq d$, for some $n > 0$ and $i < d$. |
| $(\mathbb{T}, \Pi) \models \mathbf{H}^d \neg a_\pi$ | iff | $a \notin t[p].e$ for $(t, p) = \Pi(\pi), \forall p \in \{i, ..., i+d\} \wedge (t[i+n].\tau - t[n].\tau) \geq d$, for some $n > 0$ and $i < d$ |
| $(\mathbb{T}, \Pi) \models \phi_1 \wedge \phi_2$ | iff | $((\mathbb{T}, \Pi) \models \phi_1) \wedge ((\mathbb{T}, \Pi) \models \phi_2)$ |
| $(\mathbb{T}, \Pi) \models \neg \phi$ | iff | $\neg((\mathbb{T}, \Pi) \models \phi)$ |
| $(\mathbb{T}, \Pi) \models \phi_1 \odot \phi_2$ | iff | $\exists i, j, k \text{ s.t. } i \leq k \leq j \text{ and } k = \min k' s.t. i \leq k' \leq j, (\mathbb{T}_{[i,k]}, \Pi) \models \phi_1 \wedge (\mathbb{T}_{[k+1,j]}, \Pi) \models \phi_2)$ |
| $(\mathbb{T}, \Pi) \models [\phi]^{[x,y]}$ | iff | $\exists i, j, k \text{ s.t. } k \geq i + x, (\mathbb{T}_{[k,i+y]}, \Pi) \models \phi \wedge ((\Pi)^j - (\Pi)^{now}) \geq y \text{ for some } i, j \geq 0$ |

and robustness, that can be formalized as HyperTWTL formulae as follows.

**Opacity**: In the mission described above, the private details of the user must not be compromised on any mission. Information-flow security policies define what malicious users can learn about a system while (partially) observing the system. A system is opaque if it meets two requirements: (i) there exist at least two executions of the system mapped to $\pi_1$ and $\pi_2$ with the same observations but bearing distinct secret, and (ii) the secret of each path cannot be accurately determined only by observing the system. Given a pair traces $\pi_1$ and $\pi_2$, let us assume the pick-up location $p_i$ is the only information a system user can observe, and the delivery routes are the secret to be kept from any potential malicious user. Then, opacity is guaranteed if we observe the assigned task is performed on both traces while having different routes and avoiding all obstacles from the set $O$ as well as sharing the same observations $\mathbb{B}$. This requirement can be formalized as HyperTWTL formula as:
$\varphi_{op} = \forall \pi_1 \forall \pi_2 \cdot [\mathbf{H}^1 I_{\pi_1} \wedge \mathbf{H}^1 I_{\pi_2}]^{[0, T_1]} \odot [\mathbf{H}^1 p_{i\pi_1} \wedge \mathbf{H}^1 p_{i\pi_2}]^{[T_2, T_3]} \odot [\mathbf{H}^1 d_{j\pi_1} \wedge \mathbf{H}^1 d_{j\pi_2}]^{[T_4, T_5]} \wedge [\mathbf{H}^{T_5 - T_1} \mathbb{B}_{\pi_1} \wedge \mathbf{H}^{T_5 - T_1} \mathbb{B}_{\pi_2}]^{[T_1, T_5]} \wedge [\mathbf{H}^{T_5 - T_1} \neg O_{\pi_1} \wedge \mathbf{H}^{T_5 - T_1} \neg O_{\pi_2}]^{[T_1, T_5]}$

**Side-channel attacks**: Side-channel timing attacks are usually initiated by intruders to acquire sensitive information from robotic applications by exploiting the execution time of the system. Recently, the robotic system's opacity, confidentiality, and availability have been compromised by side-channel timing attacks [2, 32]. As a countermeasure, it is required that each pair of mission executions (by a robot(s)), mapped to a pair of traces $\pi_1$ and $\pi_2$, should end up in a delivery state within close enough time after finishing their tasks while avoiding all obstacles from the set $O$. This requirement can be formalized as HyperTWTL formula as:
$\varphi_{sc} = \forall \pi_1 \forall \pi_2 \cdot [\mathbf{H}^1 I_{\pi_1} \wedge \mathbf{H}^1 I_{\pi_2}]^{[0, T_1]} \to [\mathbf{H}^1 p_{i\pi_1} \wedge \mathbf{H}^1 p_{i\pi_2}]^{[T_2, T_3]} \odot [\mathbf{H}^1 d_{j\pi_1} \wedge \mathbf{H}^1 d_{j\pi_1}]^{[T_4, T_5]} \wedge [\mathbf{H}^{T_5 - T_1} \neg O_{\pi_1} \wedge \mathbf{H}^{T_5 - T_1} \neg O_{\pi_2}]^{[T_1, T_5]}$

HyperTWTL can also be useful in expressing properties related to information-flow security, concurrency, and safety policies in various complex robotic systems [11, 12].

## 3 PROBLEM FORMULATION

We consider an agent moving over a discretized $m \times n$ environment. We model the dynamics of an agent in the given environment as a Markov Decision Process (MDP) with initially unknown probabilities. At any given state $s_i \in S$, where $i = \{0, 1, ..., m \times n - 1\}$, the set of actions $A$ that can be taken by an agent is $A =$

$\{North, East, West, South, Stay\}$. If the intended action is North ($N$), East ($E$), West ($W$), South ($S$), or Stay($St$), the agent is likely to go up, right, left, down, or stay in the current state, respectively. The transition probability between the states is denoted by $P : S \times A \times S \to [0, 1]$. We formally define an MDP and other terminologies as follows.

**Definition 1 (Markov Decision Process)**: An MDP is a tuple $\mathcal{M} = (S, s_0, A, P, l, \gamma, R)$ where $S$ is a finite set of states; $s_0 \in S$ is the initial state; $A$ is a set of actions; $P : S \times A \times S \to [0, 1]$ is a probabilistic transition relation; $R : S \times A \to \mathbb{R}$ is a reward function; $\gamma$ is the discount factor and $l : S \to 2^{AP}$ is the labeling function. A path $p$ over an MDP $\mathcal{M}$ can be defined as a sequence of states $p = s_0 \xrightarrow{a_0} s_1 \xrightarrow{a_1} s_2 \ldots$ such that every transition $s_i \xrightarrow{a_i} s_{i+1}$ is allowed in MDP $\mathcal{M}$. We assume that each state is mapped to a set of atomic propositions that hold true in that state using the labeling function $l$. Given a finite path $p = s_0 s_1 \cdots s_n$, we define a finite trace that captures the sequence of corresponding labels with assigned time-stamps as $t = (\tau_0, l(s_0))(\tau_1, l(s_1)) \cdots (\tau_n, l(s_n))$ where $n \in \mathbb{Z}_{\geq 0}$. We denote $\mathbb{T}_\mathcal{M}$ as the set of all traces generated over MDP $\mathcal{M}$. Thus, we denote the satisfaction of traces $\mathbb{T}_\mathcal{M}$ against the given formula $\varphi$ as $\mathbb{T}_\mathcal{M} \models \varphi$. In the rest of the paper, we refer to the path over a given MDP as an *episode* denoted as $ep$. In the rest of the paper, we denote the length of an episode $ep$ as $||\varphi||$ as computed in Sec. 2.1.

Given an MDP, a policy $\rho : S \to A$ is defined as stationary and deterministic if $\rho$ does not change over time and $\rho(\cdot \mid s)$ involves a single outcome, where $\rho(\cdot \mid s)$ represents a policy where $\rho$ specifies a specific action or distribution over actions to take given a particular state $s$. The agent is required to find a policy $\rho$ that maximizes the expected cumulative reward at state $s$, i.e. $\mathbb{V}_\mathcal{M}^\rho(s) = \mathbb{E}^\rho[\sum_{n=0}^{\infty} \gamma^n r(s_n, a_n) \mid s_0 = s]$, where $\mathbb{E}^\rho[\cdot]$ denotes the expected value under $\rho$, $\gamma$ is the discount factor, $n \in \mathbb{Z}_{\geq 0}$ is a time-point, $r(s_n, a_n)$ is the reward received at state $s_n$ when action $a_n$ is taken given an agent starts from state $s \in S$. Similarly, given MDP $\mathcal{M}$ we define the action-value function as $\mathbb{Q}_\mathcal{M}^\rho(s, a) = \mathbb{E}^\rho[\sum_{n=0}^{\infty} \gamma^n r(s_n, a_n) \mid s_0 = s, a_0 = a]$. We now define an optimal policy $\rho^*$ at state $s$ as $\rho^*(s) = \arg\max_{\rho \in \Gamma} \mathbb{V}_\mathcal{M}^\rho(s)$, where $\Gamma$ is a set of stationary deterministic policies.

**Definition 2 (Non-deterministic Finite Automaton)**: A Non-deterministic Finite Automaton (NFA) is a tuple $\mathcal{N} = (\mathcal{B}, b_0, \Sigma, \Delta_\mathcal{N}, F_\mathcal{N})$ where $\mathcal{B}$ is a finite set of states; $b_0 \in \mathcal{B}$ is the initial state; $\Sigma$ is a set of alphabets; $\Delta_\mathcal{N} : \mathcal{B} \times \Sigma \to \mathcal{B}$ is the transition function; and $F_\mathcal{N} \subset \mathcal{B}$ is the set of accepting states. The transition relation $\Delta_\mathcal{N}$ is nondeterministic i.e., for a state $b \in \mathcal{B}$ and symbol $a \in \Sigma$, $\Delta_{\mathcal{N}}(q, a)$ can be any subset of $\mathcal{B}$ (with the empty set included).

**Definition 3 (Deterministic Finite Automaton):** A Deterministic Finite Automaton (DFA) is a tuple $\mathcal{D} = (\mathcal{X}, x_0, \Sigma, \Delta_{\mathcal{D}}, F_{\mathcal{D}})$ where $\mathcal{X}$ is a finite set of states; $x_0 \in \mathcal{X}$ is the initial state; $\Sigma$ is a finite set of events; $\Delta_{\mathcal{D}} \subseteq \mathcal{X} \times \Sigma \times \mathcal{X}$ is the deterministic transition relation; and $F_{\mathcal{D}} \subset \mathcal{X}$ is the set of accepting states. We use $\mathcal{L}(\mathcal{D}) \subset \Sigma^*$ to denote the language of $\mathcal{D}$, i.e., the set of events in the alphabet $\Sigma$ that are accepted by $F_{\mathcal{D}}$. We use $\mathcal{L}(\mathcal{D}) \subseteq \Sigma^*$ as the language of $\mathcal{D}$, i.e., all finite traces over $\mathcal{D}$ with runs that have states ending or visiting $F_{\mathcal{D}}$. Given a set of finite traces $t_1, \cdots, t_n$ over $\mathcal{D}$, we define the point-wise product of the traces as $zip(t_1, \cdots, t_n) \in (\Sigma^n)^*$, i.e. $zip(t_1, \cdots, t_n) := (t_1, \cdots, t_n)$.

**Definition 4 (Timed Kripke Structure):** A timed Kripke structure (TKS) $\mathcal{K}$, is a tuple $\mathcal{K} = (K, k_{init}, \delta, AP, l)$ where $K$ is a finite set of states; $k_{init} \subseteq K$ is the set of initial states; $\delta \subseteq K \times \mathbb{Z}_{\geq 0} \times K$ is a set of transitions; $AP$ is a finite set of atomic propositions; and $l : K \rightarrow \Sigma$ is a labeling function on the states of $\mathcal{K}$.

**Definition 5 (Probability of Satisfying HyperTWTL):** Given a HyperTWTL formula $\varphi$, and a stationary deterministic policy $\rho$ starting from state $s$, the probability that the HyperTWTL formula $\varphi$ is satisfied across a finite set of traces $\mathbb{T}_{\mathcal{M}}$ is denoted as $P(\mathbb{T}_{\mathcal{M}} \models \varphi)$, where each $t_j \in \mathbb{T}_{\mathcal{M}}$ is given as $t_j = (\tau_0, l(s_0))(\tau_1, l(s_1)) \cdots (\tau_n, l(s_n))$, where $\tau_i \in \mathbb{R}_{\geq 0}$ is timestamp, $n \leq ||\varphi||$, $j \in [1, |\mathbb{T}_{\mathcal{M}}|]$, and $l(\cdot) : S \rightarrow \Sigma$ is the labeling function. We compute this probability over all traces induced by the stochastic dynamics of the given MDP under $\rho$. We now define the expected satisfaction probability of the given HyperTWTL formula across a set of traces as $\mathbb{E}^\rho[P(\mathbb{T}_{\mathcal{M}} \models \varphi)] = \sum_{\mathbb{T}_{\mathcal{M}}} P(\mathbb{T}_{\mathcal{M}}) \cdot P(\mathbb{T}_{\mathcal{M}} \models \varphi)$, where $P(\mathbb{T}_{\mathcal{M}})$ is the probability distribution over the trace set $\mathbb{T}_{\mathcal{M}}$ induced by $\rho$.

---

**Problem Statement:** Given an MDP $\mathcal{M}$ and a constrained mission formalized as a HyperTWTL specification $\varphi$, which is to be completed within the time-bound $||\varphi||$, the goal is to find an optimal stationary deterministic policy $\rho^*$ over multiple traces starting at state $s$ that maximizes the expected cumulative reward while incorporating the satisfaction of HyperTWTL formula $\varphi$ with a maximum probability $P$ equal to or greater than the satisfiability probability threshold $P_{th}$, i.e.,

$$\rho^*(s) = arg \; max_\rho \mathbb{E}^\rho \Big[ \sum_{n=0}^{\infty} \gamma^n r(s_n, a_n) \mid \rho \Big] \quad (2)$$

such that $arg \; max_\rho \; P(\mathcal{T}_{\mathcal{M}} \models \varphi)) \geq P_{th}$, $\forall \varphi \in \varphi_{tasks}$.

---

## 4 PROPOSED METHOD: HYPERTWTL-CONSTRAINED POLICY LEARNING

In this section, we present a solution for the problem statement presented in Section 3. Given a HyperTWTL formula $\varphi$, an MDP $\mathcal{M}$ with known states and actions, and learning parameters, our proposed policy learning approach has three parts. Given the inputs, we first generate an automaton for the HyperTWTL formula $\varphi$ by inductively constructing a $\mathcal{K}$−equivalent DFA $\mathcal{D}_{\phi_i}$ for each sub-formula $\phi_i$ from the HyperTWTL formula $\varphi$ where $i \in [1, m]$ and $m$ is the number of quantifiers in $\varphi$. We then combine the generated automaton with the input MDP $\mathcal{M}$ to generate Product MDP $\mathcal{P}$. We then use an offline algorithm, to construct the Timed MDP $\mathcal{T}$ by introducing the notion of time over the $\mathcal{P}$ and pruning $\mathcal{T}$ by removing all unaccepting states. Lastly, we use the proposed Dynamic Boltzmann Softmax-based Reinforcement Learning (hereinafter referred to as Softmax-$\epsilon$ Reinforcement Learning) in a discrete action space approach to learn and update the optimal policy that maximizes the expected rewards and also satisfies $\varphi$ using the Timed MDP $\mathcal{T}$ as the environment for the agent.

### 4.1 Automata Construction

We use a closed HyperTWTL $\varphi$ (i.e. all trace variables are quantified in $\varphi$) of the form $\varphi = Q_1 \pi_1 \dots Q_m \pi_m \cdot \psi$ where each $Q_i \in \{\forall, \exists\}$ ($i \in [1, m]$) and $\psi$ is the inner TWTL formula to specify the tasks to be performed by an agent. In this paper, we limit our formulae to alternation-free HyperTWTL formulae. Let $\phi$ be a quantified sub-formula generated from the HyperTWTL formula $\varphi$. Thus, by $\phi_i \sqsubseteq \varphi$, we mean $\phi_i$ is a sub-formula of $\varphi$. To determine the satisfiability of $\varphi$ against $\mathcal{M}$, we inductively construct the automaton $\mathcal{D}_{\phi_i}$ for each quantified sub-formulae $\phi_i \sqsubseteq \varphi$. Specifically, we first construct a Non-deterministic Finite Automaton (NFA) $\mathcal{N}_{\phi_i}$, that is $\mathcal{K} - equivalent$ to $\phi_i$ [7, 18]. The $\mathcal{K} - equivalence$ refers to the alignment of an accepted language for an automaton and the set of trace assignments that satisfy $\phi_i$ over the Kripke structure $\mathcal{K}$.

For each automaton $\mathcal{A}_{\phi_i}$ to be constructed, we define the language as $\mathcal{L}(\mathcal{A}_{\phi_i}) \subseteq (S^m)^*$, where $S^m$ is the sequence of $m$-tuples of states of $\mathcal{K}$ accepted by $\mathcal{A}_{\phi_i}$. Each m-tuple $(s_1, s_2, \dots, s_m) \in S^m$ corresponds to a simultaneous assignment of states to the $m$ trace variables $\pi_1, \pi_2, \dots, \pi_m$ in a quantified quantified sub-formulae $\phi_i \sqsubseteq \varphi$. Thus, $(S^m)^*$ denotes sequences of state tuples, which capture a synchronized execution of multiple traces, through the given Kripke structure. We also define $zip(t_1, \dots, t_n)$ as a function that interleaves the traces $t_1, \dots, t_n$ into a sequence of tuples. We can then assume that language over $\mathcal{A}_{\phi_i}$ i.e., $\mathcal{L}(\mathcal{A}_{\phi_i}) \subseteq (S^m)^*$ is $\mathcal{K} - equivalent$, if and only if for all trace assignments $\Pi$, $zip(t_1, \dots, t_n, t) \in \mathcal{L}(\mathcal{A}_{\phi_i})$, then $\Pi \models \phi_i$.

For a quantified sub-formula $\phi = \exists \pi \cdot \phi_1$, the NFA $\mathcal{N}_{\phi_i}$ is constructed by reducing the alphabet from $S^{m+1}$ (for an automaton $\mathcal{A}_{\phi_1}$) to $S^m$. This constructed NFA $\mathcal{N}_\phi$ selects potential traces by allowing transitions on all possible values for the existentially quantified trace variable $\pi$. Formally, for each transition $(q, (s_1, \dots, s_m, s_{m+1}), q') \in \Delta_{\mathcal{A}_{\phi_i}}$, we add transitions $(q, (s_1, \dots, s_m), q')$ to $\mathcal{N}_\phi$ for all possible values of $s_{m+1} \in S$. This construction ensures that $\mathcal{N}_\phi$ accepts a word $(s_1, \dots, s_m)^*$ if and only if there exists some trace assignment for $\pi$ such that the extended word is accepted by $\mathcal{A}_{\phi_1}$. For universal quantifiers $\phi = \forall \pi \cdot \phi_1$, we construct an NFA that accepts a word if and only if for all possible trace assignments to $\pi$, the extended word is accepted by $\mathcal{A}_{\phi_1}$. This is achieved by first constructing the NFA for the negation of the existential case, then complementing the output. Finally, to construct a Deterministic Finite Automaton (DFA) $\mathcal{D}_{\phi_i}$ for the subsequent construction of product automaton with the MDP, we determinize each generated NFA $\mathcal{N}_{\phi}$ into a (DFA) $\mathcal{D}_{\phi i}$ using standard subset construction [25]. We acknowledge that this construction has a high complexity due to the determinization steps. However, the focus on alternation-free fragments of HyperTWTL with small numbers of quantifiers, makes the complexity manageable. Thus, given the DFA construction $\mathcal{D}_{\phi}$, the transition function $\Delta_{\mathcal{D}_{\phi}}$ ensures that $\mathcal{D}_{\phi}$ accepts the language corresponding to the semantic meaning of $\phi$, maintaining the $\mathcal{K}$-equivalence property.

Finally, given a TKS $\mathcal{K} = (K, k_{init}, \delta, AP, l)$ and DFA $\mathcal{D}_{\phi_1} = (\mathcal{X}_1, x_0, \Sigma_1, \Delta_{\mathcal{D}_{\phi_1}}, F_{\mathcal{D}_{\phi_1}})$, we define the construction of the DFA $\mathcal{D}_{\phi} = (\mathcal{X}', x_0', \Sigma, \Delta_{\mathcal{D}_{\phi}}, F_{\mathcal{D}_{\phi}})$ as follows:

| | |
|---|---|
| $\phi = \exists \pi \cdot \phi_1$ | $\mathcal{D}_{\phi} = \mathcal{D}_{\phi_1} \times \mathcal{K} = (\mathcal{X}_1 \times K, x_0 \times k_0, \Sigma, \Delta_{\mathcal{D}_{\phi}}, F_{\mathcal{D}_{\phi_1}} \times K)$ $\Delta_{\mathcal{D}_{\phi}}((x, k), \Sigma) = ((x, k), (x', k'))$ if $x' = \Delta_{\mathcal{D}_{\phi_1}}(x, l(k)) \wedge (k, k') \in \delta$; where $\mathcal{X}' = (\mathcal{X}_1 \times K), x_0' = (x_0 \times k_0)$ and $F_{\mathcal{D}_{\phi}} = (F_{\mathcal{D}_{\phi_1}} \times K)$. |

Now, we construct a product MDP denoted as $\mathcal{P} = (S_{\mathcal{P}}, p_0, A, P_{\mathcal{P}}, l, \gamma_{\mathcal{P}}, R_{\mathcal{P}})$, from the DFA $\mathcal{D}_{\phi} = (\mathcal{X}', x_0', \Sigma, \Delta_{\mathcal{D}_{\phi}}, F_{\mathcal{D}_{\phi}})$ and the MDP $\mathcal{M} = (S, s_0, A, P, l, \gamma, R)$ where $S_{\mathcal{P}} = (S \times \mathcal{X}')$ is a finite set of states; $p_0 = (s_0, x_0')$ is the initial state; $A$ is the set of actions; $\gamma_{\mathcal{P}}$ is discount factor; $R_{\mathcal{P}} : S_{\mathcal{P}} \rightarrow \mathbb{R}$ is a reward function such that $R_{\mathcal{P}}(p) = R(s)$ for $p = (s, x') \in S_{\mathcal{P}}$; and $P_{\mathcal{P}} : S_{\mathcal{P}} \times A \times S_{\mathcal{P}} \rightarrow [0, 1]$ is a probabilistic transition relation defined as:

$$P_{\mathcal{P}}(p, a, p') = \begin{cases} P(s, a, s') & \text{if } x'' = \Delta_{\mathcal{D}_{\phi}}(x', l(s)) \\ 0 & \text{otherwise} \end{cases}$$

where $p = (s, x') \in S_{\mathcal{P}}$ and $p' = (s', x'') \in S_{\mathcal{P}}$.

**Theorem 1:** Given a HyperTWTL formula $\varphi = Q_1 \pi_1 \ldots Q_m \pi_m \cdot \psi$, let $\phi_1, \cdots, \phi_m$ be the sub-formulae generated from $\varphi$. For each $\phi_i \sqsubseteq \varphi$, if the constructed automaton $\mathcal{D}_{\phi i}$ is $\mathcal{K}$−equivalent to $\phi_i$, then $\mathbb{T}_{\mathcal{K}} \models \varphi$ iff $\mathcal{L}(\mathcal{K}) \subseteq \mathcal{L}(\mathcal{D}_{\phi i})$.

*Proof Sketch:* The proof sketch can be found in the Appendix A.

**Theorem 2:** Given a product MDP $\mathcal{P} = (S_{\mathcal{P}}, p_0, A, P_{\mathcal{P}}, l, \gamma_{\mathcal{P}}, R_{\mathcal{P}})$, from the DFA $\mathcal{D}_{\phi} = (\mathcal{X}', x_0', \Sigma, \Delta_{\mathcal{D}_{\phi}}, F_{\mathcal{D}_{\phi}})$ and the MDP $\mathcal{M} = (S, s_0, A, P, l, \gamma, R)$, for any stationary deterministic policy $\rho$, the probabilities of satisfaction of the given HyperTWTL formula $\varphi$ are preserved between $\mathcal{M}$ and $\mathcal{P}$.

*Proof Sketch:* The proof sketch can be found in the Appendix A.

## 4.2 Timed MDP Generation

It can be observed that the constructed MDP $\mathcal{P}$ for the HyperTWTL $\varphi$ does not contain any notion of time that is required to determine the satisfiability of any mission requirement formalized as a Hyper-TWTL specification. To address this, we construct a Timed MDP for tracking the time progression in any policy search given as HyperTWTL specifications. We now formally define a Timed MDP and other important parameters to successfully construct Timed MDPs given a HyperTWTL formula $\varphi$ and a standard MDP $\mathcal{M}$ as follows:

**Definition 6 (Timed MDP):** Given the product MDP $\mathcal{P} = \{S_{\mathcal{P}}, p_0, A, \Delta_{\mathcal{P}}, R_{\mathcal{P}}, F_{\mathcal{P}}\}$ and a time set $\mathbf{T} = \{0, \cdots, n, \cdots, ||\varphi|| - 1\}$, a Timed MDP $\mathcal{T}$ is a tuple $\mathcal{T} = (Q, q_0, A, \Delta_{\mathcal{T}}, R_{\mathcal{T}}, F_{\mathcal{T}})$ where, $Q = S_{\mathcal{P}} \times \mathbf{T}$ is a set of finite states, with each state $q_n \in Q$ represented as a pair $(p_n, n)$ where $p_n \in S_{\mathcal{P}}$ and $n \in \mathbf{T}$; $q_0 = \{(p_0, 0)\}$ is a set of initial states of $\mathcal{T}$, where path starts from state $p_0$ at time-point $n = 0$; $A$ is a finite set of actions; $R_{\mathcal{T}} : Q \rightarrow \mathbb{R}$ is the reward function such that $R_{\mathcal{T}}(q_n) = R_{\mathcal{P}}(p_n)$ for $q_n \in Q$; $F_{\mathcal{T}} = (F_{\mathcal{P}} \times \mathbf{T}) \subseteq Q$ is a set of accepting states; $\Delta_{\mathcal{T}} : Q \times A \times Q \rightarrow [0, 1]$ is the probabilistic transition relation defined as

$$\Delta_{\mathcal{T}}(q_n, a, q_{n+1}) = \begin{cases} P_{\mathcal{P}}(p, a, p') & \text{if } n + 1 < ||\varphi|| \\ 0 & \text{otherwise}; \end{cases}$$

To guarantee that time progresses steadily until the horizon is reached, transitions remain fixed at the final step ($n = ||\varphi|| - 1$), i.e., $\Delta_{\mathcal{T}}(q_n, a, q_{n+1}) = P_{\mathcal{P}}(p, a, p')$.

Given a Timed MDP $\mathcal{T}$, for any state $q_n \in Q$, we define the set of reachable states under action $a$ as

$$Q^{reach}(q_n, a) = \begin{cases} \{(p', n+1)|\delta_{\mathcal{P}}(p, a, p') > 0\} & \text{if } n + 1 < ||\varphi|| \\ \{(p', n)|\delta_{\mathcal{P}}(p, a, p') > 0\} & \text{if } n = ||\varphi|| - 1 \end{cases}$$

**Definition 7 ($\varepsilon$−Probabilistic):** Given any Timed MDP $\mathcal{T}$ and an $\varepsilon \in [0, 1)$, we conclude the transition $(q_n, a, q_{n+1})$ exhibits $\varepsilon$−probabilistic characteristics if the probability associated with the transition is at least $1 - \varepsilon$, i.e., $\Delta_{\mathcal{T}}(q_n, a, q_{n+1}) \geq 1 - \varepsilon$. From the above definition, we can conclude that as $\varepsilon$ approaches 1, any transition $(q_n, a, q_{n+1})$ exhibits $\varepsilon$-probabilisticity [1]. Now, given any two states, $(q_n, q_{n+i}) \in Q$ and $\varepsilon \in [0, 1)$, we define the distance under $\varepsilon$−Probabilistic transitions denoted as $dist^{\varepsilon}(q_n, q_{n+i}) = i$ as the distance between the two states under $\varepsilon$−Probabilistic transitions if there exists a path from $q_n$ to $q_{n+i}$ under $\varepsilon$−Probabilistic transitions [27].

**Definition 8 (Distance to $F_{\mathcal{T}}$):** Given a state $q_n \in Q$, the distance to $\mathcal{D}_{\mathcal{T}}$ denoted as $F_{\mathcal{T}}^{dist}$ under $\varepsilon$−probabilistic transitions can be computed as $F_{\mathcal{T}}^{dist}(q_n) = \min_{q_{n+i} \in Q} dist^{\varepsilon}(q_n, q_{n+i})$.

**Definition 9 ($F_{\mathcal{T}}^{reach}$-Policy):** For any given Timed MDP $\mathcal{T}$ and $\varepsilon \in [0, 1)$, we define a stationary policy to be generated to reach $F_{\mathcal{T}}$ over a Timed MDP denoted as $F_{\mathcal{T}}^{reach} : Q \rightarrow A$ as $F_{\mathcal{T}}^{reach}(q_n) = \arg \min_{a \in A} F_{\mathcal{T}}^{dist}(q_n, a)$, where $F_{\mathcal{T}}^{dist}(q_n, a)$ is the smallest distance to $\mathcal{D}_{\mathcal{T}}$ among the states reachable from $q_n$ with a probability of at least $1 - \varepsilon$ for any given state $q_n \in Q$ and $i \geq 0$. Under the policy $F_{\mathcal{T}}^{reach}(q_n)$, we define the probability of reaching $\mathcal{D}_{\mathcal{T}} \in \mathcal{T}$ from $q_n$ in the next $i \geq 0$ time steps as $P(q_n \xrightarrow{i} F_{\mathcal{T}}; F_{\mathcal{T}}^{reach}(q_n))$. For any given state $q_n \in Q$, if $F_{\mathcal{T}}^{dist}(q_n) < \infty$, then we conclude that

$$P(q_n \xrightarrow{i} F_{\mathcal{T}}; F_{\mathcal{T}}^{reach}(q_n)) \geq \sum_{j=0}^{\frac{i - F_{\mathcal{T}}^{dist}(q_n)}{2}} \frac{i}{(i-1)!j!} \varepsilon^j (1-\varepsilon)^{i-j}, \quad (3)$$

such that $i \geq F_{\mathcal{T}}^{dist}(q_n)$ [1]. However, it is noteworthy that when a maximum of unintended transitions (i.e., transitions that increase the distance to the set of accepting states by at most one with

---

**Algorithm 1:** Offline construction of the pruned Timed MDP

---

**Inputs** : HyperTWTL formula ($\varphi$), MDP ($\mathcal{M}$), Episode length ($||\varphi||$), Satisfaction probability threshold ($P_{th}$), Estimated motion uncertainty ($\varepsilon$)

**Outputs** : Pruned Timed MDP ($\mathcal{T}$)

1: Generate sub-formulae $\phi_1, \cdots, \phi_m$ from $\varphi$ for $m$ quantifiers
2: Construct $\mathcal{K} - equivalent$ DFA for $\phi_i \sqsubseteq \phi$, $\mathcal{D}_{\phi_i} = (\mathcal{B}, b_0, \Sigma, \mathcal{D}, F_\mathcal{D})$
3: Create Product MDP, $\mathcal{P} = (\mathcal{D}_{\phi_i} \times \mathcal{M}) = (S_\mathcal{P}, p_0, A, \Delta_\mathcal{P}, R_\mathcal{P}, F_\mathcal{P})$
4: Create Timed MDP, $\mathcal{T} = \mathcal{P} \times \{0, \cdots, n, \cdots ||\varphi|| - 1\} = (Q, q_0, A, \Delta_\mathcal{T}, R_\mathcal{T}, F_\mathcal{T})$
5: Compute $F_\mathcal{T}^{dist}(q_n)$ for each $q_n \in Q$
6: **Initialization:** $A_{Fes}(q_n) = \{A | q_n \in Q\}$
7: **while** $q_n \notin F_\mathcal{T}$ **do**
8:   **for** each action $a \in A_{Fes}(q_n)$ **do**
9:     Compute $Q^{reach}(q_n, a)$
10:     $dist_{max} = \max\limits_{q_{n+1} \in Q^{reach}(q_n,a)} F_\mathcal{T}^{dist}(q_{n+1});$
11:     $i = ||\varphi|| - n - 1$ (the remaining episode time);
12:     $u_{max} = \left\lfloor \frac{i-1-dist_{max}}{2} \right\rfloor;$
13:     **if** ($u_{max} \geq 0$) and $\sum_{j=0}^{u_{max}} \frac{i!}{(i-j)!j!} \varepsilon^j (1-\varepsilon)^{i-1-j} < P_{th}$ or ($u_{max} < 0$) **then**
14:       $A_{Fes}(q_n) = A_{Fes}(q_n) \setminus \{a\};$
15:     **end if**
16:   **end for**
17: **end while**
18: **return** $\mathcal{T}$

---

the remaining probability from $1 - \varepsilon$) denoted as $u_{max}$ are observed within $i \geq u_{max}$ transitions, the given $\mathcal{T}$ reaches $F_\mathcal{T}$ iff $u_{max} \leq i - u_{max} - F_\mathcal{T}^{dist}(q_n)$, i.e., $u_{max} \leq \frac{i - F_\mathcal{T}^{dist}(q_n)}{2}$. For any $j \leq u_{max}$, the number of all possible $i$-length sequences of $j$ unintended transitions and $i - j$ intended transitions (i.e., state transitions that reduce the distance to the accepting states by one with probability of at least $1 - \varepsilon$) is the $\frac{i!}{(i-j)!}$. Recall that, every intended transition occurs with a probability of at least $1 - \varepsilon$, for any $i \geq u_{max}$ the minimum bound of probability while observing a sequence of $i$ transitions involving at least $i - u_{max}$ intended transitions is given as $\sum_{j=0}^{u_{max}} \frac{i!}{(i-j)!i!} \varepsilon^i (1-\varepsilon)^{i-j}$.

Based on the above, we derive Algorithm 1 inspired by [1] for the construction of Timed MDP $\mathcal{T}$ and pruning of the feasible actions at each $q_n \in Q$. Specifically, once the Timed MDP $\mathcal{T}$ is constructed, we prune $\mathcal{T}$ by removing all sequence of states which do not eventually end in $F_{\mathcal{D}_\phi}$. In summary, any action at any given state $q_n$ is removed if (1) Equation (3) does not hold for any transition into state $q_{n+1}$, or (2) the remaining episode time is smaller than $F_\mathcal{T}^{dist}(q_{n+1})$. Through pruning, we guarantee that the satisfaction of the HyperTWTL formula $\varphi$ can be determined within the time horizon $||\varphi||$, given the remaining states in $\mathcal{T}$, thus effectively reducing the state space only states that are reachable.

Algorithm 1 is an offline construction of Timed MDP $\mathcal{T}$ for a given task expressed as a HyperTWTL specification. We use the proposed Algorithm 1 to quantify the worst-case probability of constraint satisfaction in the remaining $i$ time steps from a given state $q_n$ of a Timed MDP $\mathcal{T}$. For any given feasible action set of $q_n$, an action $a$ is removed if the worst-case satisfaction probability of any state in the set of potential next states is less than the desired probability. The Algorithm takes as inputs the tasks specified as a HyperTWTL formula $\varphi$, the MDP ($\mathcal{M}$), the satisfaction probability threshold $P_{th}$, the length of each episode $||\varphi||$ computed from the time bound of $\varphi$, and the algorithm parameter $\varepsilon$. The Algorithm first

---

**Algorithm 2:** Softmax-$\varepsilon$ reinforcement learning for Hyper-TWTL Specifications

---

**Inputs** : Pruned Timed MDP ($\mathcal{T}$)

**Outputs** : Policy ($\pi$)

1: **Initialization:** Policy parameters $\sigma$, $\varepsilon$, $\theta$
2: **Initialization:** Initial Q-table
3: Assign task specified as $\varphi$
4: **for** $ep = 1 : N_{ep}$ **do**
5:   $q_n = (q_0, 0);$
6:   **for** $n = 0 : ||\varphi|| - 1$ **do**
7:     **while** $q_n \notin F_\mathcal{T}$ **do**
8:       Compute Q-values for all actions of $q_n$:
        $\mathbb{Q}_\mathcal{M}^\rho(q_n, a) = \sum_{n=0}^{||\varphi||-1} (r_n + \gamma max \cdot \mathbb{Q}_\mathcal{M}^\rho(q_{n+1}, a))$
9:       **if** $\alpha > \varepsilon$ **then**
10:         Select an action $a_n$ via $\varepsilon$-greedy
11:       **else**
12:         $a_n \leftarrow \rho(a|q_n, \theta) = \frac{exp(Q(q_n,a)/\sigma)}{\sum_{a'} exp(Q(q_n,a')/\sigma)}$
13:       **end if**
14:       Take action $a_n$ in the environment, observe $r_n$ and next state $q_{n+1}$
15:       Store the experience $(q_n, a_n, r_n, q_{n+1})$ for future updates
16:     **end while**
17:   **end for**
18: **end for**

---

generates sub-formulae $\phi_i$ from the given HyperTWTL formula $\varphi$ consistent with the number of quantifiers $m$ (Line 1). For each sub-formula $\phi_i \sqsubseteq \varphi$, we construct an DFA $\mathcal{D}_{\phi_i}$ that is $\mathcal{K} - equivalent$ to $\phi_i$ (Line 2). Subsequently, we then construct the MDP $\mathcal{P}$, the Timed MDP $\mathcal{T}$, and the $F_\mathcal{T}^{dist}(q_n)$ (Lines 3-5). The set of feasible actions $A_{Fes}(\cdot)$ for each state of the Timed MDP is then initialized with the set of actions of the MDP, $A$ (Line 6). Considering some actions in $A$ at particular states do not lead to the satisfaction of the input formula $\varphi$, the algorithm aims to prune the action sets to ensure the probabilistic satisfaction of $\varphi$. At each non-accepting state $q_n$ and each action $a$ that can be taken at $q_n$, we first generate the set of states $Q^{reach}(q_n, a)$, that can be reached from $q_n$ under action $a$ (Lines 7-9). The maximum distance to $F_\mathcal{T}$ from ($dist_{max}$) is then computed (Line 10). $dist_{max}$ allows us to compute the remaining $k$ number of actions to be taken within the remaining episode time. We then calculate $u_{max}$ based on $i$ and $dist_{max}$ (Line 12). If $u_{max} \geq 0$, then the $F_\mathcal{T}^{dist}(q_n)$ is less than the number of actions that can be taken in the next time step $i - 1$ while incorporating the scaling function $f(\cdot)$ (Line 13). If $u_{max} < 0$, then $a$ is pruned from the $A_{Fes}(q_n)$ (Line 14).

### 4.3 Policy Learning via Dynamic Boltzmann Softmax-based Reinforcement Learning

In this paper, we leverage the Boltzmann softmax and $\varepsilon$-greedy strategy to learn and update the optimal policy that maximizes the expected cumulative reward. Given the action-value function $\mathbb{Q}_\mathcal{M}^\rho(s, a)$, state $s$, action set $A$ and the epsilon parameter $\varepsilon$, the epsilon-greedy policy $\rho(s)$ is defined as

$$\rho(s) = \begin{cases} random\ action\ a \in A, & with\ probability\ \varepsilon \\ \arg\max\limits_{a \in A} \mathbb{Q}_\mathcal{M}^\rho(s, a) & with\ probability\ 1 - \varepsilon \end{cases} \quad (4)$$

Assuming a parameterized policy denoted as $\rho(a \mid s, \theta)$, represents the probability of taking action $a$ at a given state $s$ with the policy parameters $\theta$, the policy update equation using the softmax equation is given as $\Delta(\theta) = \mathbb{E}[\Sigma_a \Delta\rho(a \mid s, \theta) \mathbb{Q}_\mathcal{M}^\rho(s, a)]$, where $\Delta(\theta)$ is the gradient of the expected return with respect to $\theta$ and $\rho(a \mid s, \theta)$ is the softmax policy that determines the probability of taking action $a$ in state $s$ according to $\theta$. We formally define

$$\rho(a|s, \theta) = \frac{exp \frac{Q^\rho_M(s,a)}{\sigma}}{\sum_{a'} exp \frac{Q^\rho_M(s,a')}{\sigma}}, \quad (5)$$

where $\sigma$ is the temperature parameter controlling the exploration-exploitation trade-off.

Finally, we propose the online Algorithm 2, where we combine $\varepsilon$-greedy and softmax to select actions based on a stochastic policy. The algorithm takes as input the Timed MDP $\mathcal{T}$ generated from Algorithm 1 and returns a policy that satisfies the given HyperTWTL formula $\varphi$. We initialize policy parameters and the initial Q-table required to learn the optimal policy (Lines 1-2). We then assign a task formalized as a HyperTWTL specification to each episode $ep$, to guide the policy search (Line 3). For each $ep$, we initialize the policy to start from the specified initial state (Line 4). At any given Timed MDP $\mathcal{T}$ state $q_n$, we include the satisfaction of HyperTWTL formula $\varphi$, in the computation of Q-values for all actions (Line 8). The computed values represent the expected rewards for taking all actions in the state $q_n$. We dynamically generate a random number $\alpha \in [0, 1]$ to provide a randomized mechanism to take either greedy or non-greedy actions. If $\alpha > \varepsilon$, then an action is selected via $\varepsilon$-greedy approach otherwise compute the action probabilities using the softmax function in Equation 5 and select an action based on the probability distribution $\rho(a|q_n, \theta)$ (Lines 9-12). We observe the environment and store the experience based on the actions taken, rewards given, and the next state for later updates (Lines 13 -17).

**Theorem 3:** Let $\varphi$ be the HyperTWTL formula to be satisfied with a probability of at least $P_{th}$ within the time bound $||\varphi||$. Given an MDP $\mathcal{M}$, Timed MDP $\mathcal{T}$ and some $\varepsilon \in [0, 1)$, we assume each transition over $\mathcal{M}$ exhibits $\varepsilon$−probabilisticity, while any feasible transition over $\mathcal{T}$ can potentially increase the Distance to $F_{\mathcal{T}}$) by a maximum of one. Given the set of initial states $q_0$ of $\mathcal{T}$ satisfies $P_{th} \leq \sum_{j=0}^{\frac{||\varphi||-F_{\mathcal{T}}^{dist}(q)}{2}} \frac{||\varphi||}{(||\varphi||-1)!j!} \varepsilon^j (1-\varepsilon)^{||\varphi||-j}, \forall q \in q_0$, then $P(t_1^j, t_2^j, \cdots, t_m^j \models \varphi) \geq P_{th}, \forall j \geq 0$, where $t_1^j, t_2^j, \cdots, t_m^j$ are traces generated from episode $ep^j$, $F_{\mathcal{T}}^{dist}(q)$ represents the minimum distance from state $q$ to any accepting state in $\mathcal{D}_{\mathcal{T}}$ under $\varepsilon$-probabilistic transitions, each trace $t_i^j$ represents a path over the MDP following policy $F_{\mathcal{T}}^{reach}$, and $m$ is the number of trace variables in the given HyperTWTL formula $\varphi$.

*Proof sketch:* The proof sketch can be found in the Appendix A.

# 5 IMPLEMENTATION AND SIMULATIONS

In this section, we evaluate the feasibility of our proposed approach for learning an optimal policy for monitoring the pick-up and delivery mission presented in Section 2.2. Recall that on each mission, delivery drones are required to perform a pick-up task within the time bound $[0, T_1]$ followed by a delivery task within the time bound $[T_2, T_3]$ while guaranteeing opacity and countermeasures against side-channel attack objectives. Based on the two pick-up and two delivery locations, we define 8 different sets of pick-up and delivery missions, i.e., $\varphi_{all} = \{\varphi_{task_{s_1}}, \ldots, \varphi_{task_{s_8}}\}$.

**Table 2: Simulation parameters for comparison of Softmax-$\varepsilon$, Dyna-Q, and Q-learning**

| $P_{th}$ | $\varepsilon$ | $\varepsilon$-greedy | $\sigma$ | $N_{samples}$ | $N_{ep}$ |
|---|---|---|---|---|---|
| 0.85 | 0.05 | 0.1 | 1.0 | 20 | 50000 |

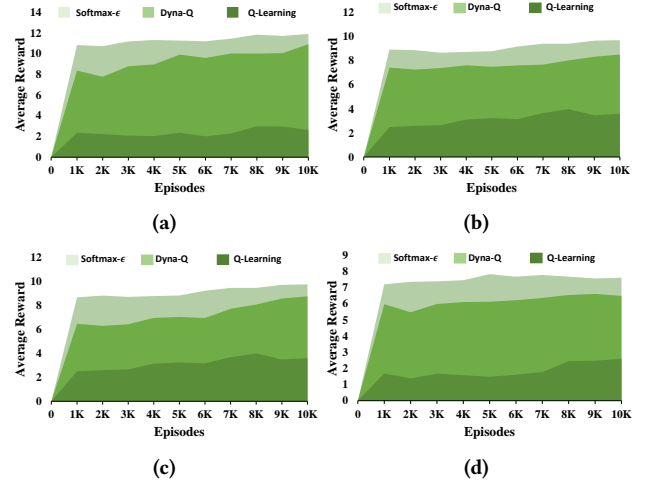

**Figure 2: Comparison of Softmax-$\varepsilon$, Dyna-Q, and Q-learning for $\varphi_{op_3}$. Task: (a) $\varphi_{p_1d_1}$, (b) $\varphi_{p_1d_2}$, (c) $\varphi_{p_2d_1}$, and (d) $\varphi_{p_2d_2}$**

Note, $\varphi_{task_{s_1}} - \varphi_{task_{s_4}}$ are $\varphi_{op}$ specifications, and $\varphi_{task_{s_5}} - \varphi_{task_{s_8}}$ are $\varphi_{sc}$ specifications. Thus, $\varphi_{all} = \{\varphi_{op_1}, \ldots, \varphi_{op_4}, \varphi_{sc_1}, \ldots, \varphi_{sc_4}\}$. Recall, each $\varphi_{task_i} \in \varphi_{all}$ contains four pick-up and delivery tasks, i.e., $\{\varphi_{p_1d_1}, \varphi_{p_1d_2}, \varphi_{p_2d_1}, \varphi_{p_2d_2}\}$. For any given task, $\varphi_{p_id_j}$ denotes a pick-up and delivery mission where the pick-up state is $p_i$ and the delivery state is $d_j$.

We conduct simulations to compare the performance between the Softmax-$\varepsilon$ approach and the Dyna-Q approach proposed in [6] as well as the Softmax-$\varepsilon$ and Q-learning approaches. Similar to [6], we assess the impact of action uncertainty on the reward optimization and scalability of the algorithm. All simulations were performed in Python 2.7 on a Windows 10 system with 16 GB RAM and Intel Core(TM) i7-4790 CPU @ 3.60 GHz. Throughout the simulations, the following time bounds are considered: $T_1 = 5$, $T_2 = 6$ $T_3 = 20$, $T_4 = 21$, $T_5 = 35$.

## 5.1 Experimental Results

To compare the sample efficiency between our approach, Dyna-Q and Q-learning, over a given number of episodes $N_{ep}$, we define a measure $\mathcal{E}_{@ep}$ as:

$$\mathcal{E}_{@ep} = \frac{\text{episode reward using Softmax-}\varepsilon}{\text{episode reward using Dyna-Q/Q-learning}} \quad (6)$$

If $\mathcal{E}_{@ep} = 1$, then both Softmax-$\varepsilon$ and Dyna-Q or Softmax-$\varepsilon$ and Q-learning exhibit the same sample efficiency over the number of episodes $N_{ep}$. However, if $\mathcal{E}_{@ep} > 1$, then Softmax-$\varepsilon$ performs better. For each $\varphi_{task} \in \varphi_{all}$, we implement both Dyna-Q, Q-learning, and Softmax-$\varepsilon$ reinforcement learning algorithms for over 50,000 episodes. The important parameters used in the simulations are shown in Table 2.

**Table 3: $\mathcal{E}_{@3\times10^4}$ values for $\varphi_{all}$ for Softmax-$\varepsilon$ and Dyna-Q (DQ) and Softmax-$\varepsilon$ and Q-learning (QL)**

|  | $\varphi_{op_1}$ | | $\varphi_{op_2}$ | | $\varphi_{op_3}$ | | $\varphi_{op_4}$ | | $\varphi_{sc_1}$ | | $\varphi_{sc_2}$ | | $\varphi_{sc_3}$ | | $\varphi_{sc_4}$ | |
|---|---|---|---|---|---|---|---|---|---|---|---|---|---|---|---|---|
|  | DQ | QL | DQ | QL | DQ | QL | DQ | QL | DQ | QL | DQ | QL | DQ | QL | DQ | QL |
| $\varphi_{p_1d_1}$ | 1.05 | 1.23 | 1.03 | 1.24 | 1.03 | 1.19 | 1.03 | 1.22 | 1.04 | 1.23 | 1.02 | 1.20 | 1.05 | 1.22 | 1.02 | 1.22 |
| $\varphi_{p_1d_2}$ | 1.02 | 1.24 | 1.03 | 1.24 | 1.03 | 1.20 | 1.03 | 1.22 | 1.02 | 1.21 | 1.03 | 1.22 | 1.02 | 1.19 | 1.03 | 1.19 |
| $\varphi_{p_2d_1}$ | 1.03 | 1.23 | 1.05 | 1.26 | 1.03 | 1.20 | 1.03 | 1.20 | 1.03 | 1.22 | 1.02 | 1.22 | 1.03 | 1.20 | 1.04 | 1.20 |
| $\varphi_{p_2d_2}$ | 1.03 | 1.22 | 1.03 | 1.24 | 1.04 | 1.22 | 1.02 | 1.21 | 1.02 | 1.21 | 1.02 | 1.22 | 1.04 | 1.20 | 1.04 | 1.22 |

**Table 4: Scalability of Softmax-$\varepsilon$ RL algorithm**

| Grid size | Number of samples | Number of episodes ($N_{ep}$) | Number of tasks | Episode length $\|\varphi\|$ | Execution time (s) |
|---|---|---|---|---|---|
| $20^2$ | 10 | 100000 | 20 | 43 | 424.82 |
| $40^2$ | 10 | 100000 | 20 | 60 | 1172.02 |
| $60^2$ | 10 | 100000 | 20 | 80 | 1752.53 |
| $80^2$ | 10 | 100000 | 20 | 100 | 3817.68 |
| $100^2$ | 10 | 100000 | 20 | 120 | 7221.12 |

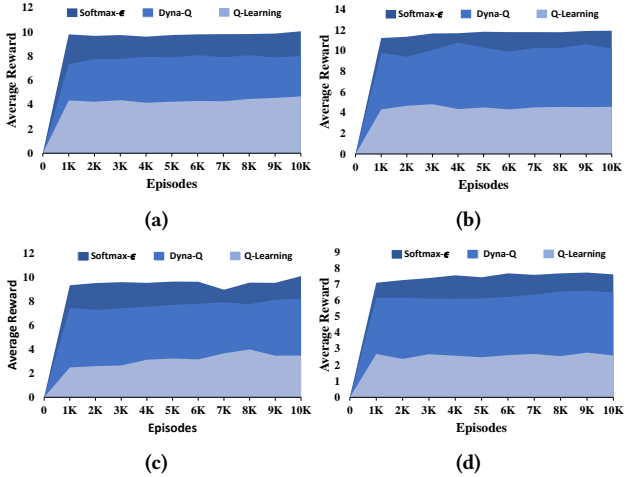

**Figure 3: Comparison of Softmax-$\varepsilon$, Dyna-Q, and Q-learning for $\varphi_{sc_2}$. Mission: (a) $\varphi_{p_1d_1}$, (b) $\varphi_{p_1d_2}$, (c) $\varphi_{p_2d_1}$, and (d) $\varphi_{p_2d_2}$**

The obtained results from the experiments are presented in Table 3. From Table 3, the results show that in learning policies for the pick-up and delivery missions while guaranteeing both opacity and countermeasures against side-channel attacks, Softmax-$\varepsilon$ performs better than Dyna-Q. For instance, for $\varphi_{op_1}$, the sample efficiencies for $\varphi_{p_1d_1}$, $\varphi_{p_1d_2}$, $\varphi_{p_2d_1}$ and $\varphi_{p_2d_2}$ are 1.05, 1.02, 1.03, and 1.03 respectively. Again, for $\varphi_{op_3}$, the respective sample efficiencies for $\varphi_{p_1d_1}$, $\varphi_{p_1d_2}$, $\varphi_{p_2d_1}$ and $\varphi_{p_2d_2}$ are 1.03, 1.03, 1.03, and 1.04. Similarly, for $\varphi_{sc_2}$, the sample efficiencies at $\mathcal{E}_{@1\times10^6}$ for $\varphi_{p_1d_1}$, $\varphi_{p_1d_2}$, $\varphi_{p_2d_1}$ and $\varphi_{p_2d_2}$ are 1.02, 1.03, 1.02, and 1.02. Once again, the sample efficiencies for $\varphi_{p_1d_1}$, $\varphi_{p_1d_2}$, $\varphi_{p_2d_1}$ and $\varphi_{p_2d_2}$ for $\varphi_{sc_4}$ are 1.02, 1.03, 1.04, and 1.04. From the results, we observe that our proposed Softmax-$\varepsilon$ RL algorithm performs better than Dyna-Q in all scenarios. In Figure 2, we observe that the sample efficiency of Softmax-$\varepsilon$ is higher than the sample efficiency of Dyna-Q algorithms for all tasks, i.e., $\varphi_{p_1d_1}$, $\varphi_{p_1d_2}$, $\varphi_{p_2d_1}$ and $\varphi_{p_2d_2}$ of $\varphi_{op_3}$. Similarly, from Figure 3 we observe again that the Softmax-$\varepsilon$ algorithm exhibits higher sample efficiency than that of the Dyna-Q algorithm while monitoring tasks, $\varphi_{p_1d_1}$, $\varphi_{p_1d_2}$, $\varphi_{p_2d_1}$ and $\varphi_{p_2d_2}$ for $\varphi_{sc_2}$. Similarly, from Table 3, the results show that in learning policies for the pick-up and delivery missions, while guaranteeing both opacity and side-channel attacks, Softmax-$\varepsilon$ outperforms the Q-learning algorithm. For instance, for $\varphi_{op_1}$, the sample efficiencies for $\varphi_{p_1d_1}$, $\varphi_{p_1d_2}$, $\varphi_{p_2d_1}$ and $\varphi_{p_2d_2}$ are 1.23, 1.24, 1.23, and 1.22 respectively. Again, for $\varphi_{op_3}$, the respective sample efficiencies for $\varphi_{p_1d_1}$, $\varphi_{p_1d_2}$, $\varphi_{p_2d_1}$ and $\varphi_{p_2d_2}$ are 1.19, 1.20, 1.20, and 1.21. Similarly, for $\varphi_{sc_1}$, the sample efficiencies at $\mathcal{E}_{@1\times10^6}$ for $\varphi_{p_1d_1}$, $\varphi_{p_1d_2}$, $\varphi_{p_2d_1}$ and $\varphi_{p_2d_2}$ are 1.23, 1.21, 1.22, and 1.21. Once again, the sample efficiencies for $\varphi_{p_1d_1}$, $\varphi_{p_1d_2}$, $\varphi_{p_2d_1}$ and $\varphi_{p_2d_2}$ for $\varphi_{sc_4}$ are 1.22, 1.19, 1.20, and 1.22.

## 5.2 Scalability Analysis

In the second set of experiments, we investigate the scalability of the proposed Softmax-$\varepsilon$ RL algorithm. We randomly generated 20 sets of pick-up and delivery missions, of which 10 are opacity-aware specifications ($\varphi_{op}$) and the other 10 are side-channel attack specifications ($\varphi_{sc}$). While keeping the number of samples, number of episodes, and number of tasks fixed, we vary the grid size from $20^2$ to $100^2$ and the length of each episode from $\|\varphi\|$=43 to $\|\varphi\|$ = 120. We then analyze the impact of grid size on the performance of the proposed approach. The results obtained with other important parameters are shown in Table 4. We observe from Table 4 that for a grid size of $20^2$ and $\|\varphi\|$=43, the Softmax-$\varepsilon$ algorithm takes 424.82 seconds to learn optimal control policies. Similarly, the time taken to learn a policy increases to 1172.02 seconds when the grid size and the length of episode increase to $40^2$ and $\|\varphi\|$ = 60, respectively. With a grid size of $100^2$ and an episode length of $\|\varphi\|$= 120, the time our proposed algorithm takes to learn a policy increases to 7221.12 seconds. From the results shown, we observe a linear trend for the time taken to learn an optimal policy using the Softmax-$\varepsilon$ RL algorithm.

## 6 RELATED WORKS

Model-free reinforcement learning algorithms have been extensively used with temporal logic to learn and synthesize optimal policies while ensuring compliance with expected behavior and constraints. In [3, 14, 22–24, 34, 35], LTL was formally used to express high-level complex tasks. The authors then introduced advanced RL algorithms that incorporate reward-sharing, safety value functions, and quantum action selection in learning optimal policies in various applications. Reinforcement learning algorithms have also been used with STL in [15, 30, 36, 38] to learn policies that guarantee safety and security in complex continuous dynamical systems. Similarly, MTL with finite time constraints has been used with reinforcement learning to design learning frameworks for robotic task planning, runtime monitoring, and self-correction problems in [28, 31]. TWTL with reinforcement learning has recently received attention from researchers. In [5], the authors used TWTL to guide the learning of policies that maximize the expected sum of rewards in unknown and unpredictable environments. Consequently, in [6], the authors proposed the Dyna-Q RL algorithm with TWTL to address the limitations of the algorithm in [5] to improve the process of maximizing the expected sum of rewards. However, to the best of our knowledge, this work will be the first to use HyperTWTL to guide the learning of policies using RL in safety- and privacy-critical robotic applications.

## 7 CONCLUSION

In this paper, we proposed a hyper-temporal logic-constrained Dynamic Boltzmann Softmax Reinforcement Learning for learning optimal policies that maximize the expected sum of rewards under unknown environments with stochastic uncertainties in bounded missions. We modeled the agent dynamics as a Markov Decision Process with initial unknown transition probabilities, while the bounded tasks were expressed as HyperTWTL specifications. Specifically, the proposed approach uses the Boltzmann softmax approach with $\varepsilon$-greedy strategy to introduce a more adaptive and sensitive exploration for the random selection of actions. We demonstrated our approach's feasibility, performance, and scalability using a pickup and delivery case study and compared the results with other baseline RL algorithms. In the future, we plan to extend our proposed approach for $k$-alternation and asynchronous fragments of HyperTWTL specifications.

# APPENDIX

## A. Theorem Proof Sketches

**Theorem 1:** Given a HyperTWTL formula $\varphi = Q_1 \pi_1 \dots Q_m \pi_m \cdot \psi$, let $\phi_1, \dots, \phi_m$ be the sub-formulae generated from $\varphi$. For each $\phi_i \sqsubseteq \varphi$, if the constructed automaton $\mathcal{D}_{\phi_i}$ is $\mathcal{K} - equivalent$ to $\phi_i$, then $\mathbb{T}_{\mathcal{K}} \models \varphi$ iff $\mathcal{L}(\mathcal{K}) \subseteq \mathcal{L}(\mathcal{D}_{\phi_i})$.

*Proof Sketch:* To check if $\mathcal{K} \models \varphi$, we inductively construct an automata that is $\mathcal{K}$-equivalent to $\phi_i$ for each sub-formula $\phi_i \sqsubseteq \varphi$. The automata $\mathcal{D}_{\phi_i}$ accepts exactly the models that satisfy $\phi_i$. Thus, the automaton $\mathcal{D}_\varphi$ accepts models satisfying the conjunction of all sub-formulae, $\phi_i, \phi_{i+1}, \dots, \phi_m$. By construction, $\mathcal{L}(\mathcal{D}_\varphi) = \mathcal{L}(\mathcal{D}_{\phi_i}) \cap \mathcal{L}(\mathcal{D}_{\phi_{i+1}}) \cap \dots \cap \mathcal{L}(\mathcal{D}_{\phi_m})$. We preserve automata equivalence by ensuring all accepting states for all $\mathcal{D}_{\phi_i}$ are in $\varphi$ as well as preserving the notion of occurrence of the propositions on any pair of traces. Let us recall that, for each sub-formula $\phi_i$, the corresponding automaton over $\Sigma$ is $\mathcal{K}$-equivalent to $\phi_i$ if for all traces, $zip(t_1, \dots, t_n) \in \mathcal{L}(\mathcal{D}_{\phi_i})$. Given that the equivalence of two systems is defined by the acceptance of the same set of strings over an input set, constructing the automaton $\mathcal{D}_\varphi$ ensures that each sub-formula $\phi_i$ accepts the models that satisfy $\varphi$ over $\mathcal{K}$.

**Theorem 2:** Given a product MDP $\mathcal{P} = (S_\mathcal{P}, p_0, A, P_\mathcal{P}, l, \gamma_\mathcal{P}, R_\mathcal{P})$, from the DFA $\mathcal{D}_\phi = (\mathcal{X}', x_0', \Sigma, \Delta_{\mathcal{D}_\phi}, F_{\mathcal{D}_\phi})$ and the MDP $\mathcal{M} = (S, s_0, A, P, l, \gamma, R)$, for any stationary deterministic policy $\rho$, the probabilities of satisfaction of the given HyperTWTL formula $\varphi$ are preserved between $\mathcal{M}$ and $\mathcal{P}$.

*Proof Sketch:* For a given product MDP $\mathcal{P}$, each state $p = (s, x')\in S_\mathcal{P}$ is expressed as a pair of state $s \in S$ from MDP $\mathcal{M}$ and state $x' \in \mathcal{X}'$ from DFA $\mathcal{D}_\phi$. For any action $a \in A$, a transition in $\mathcal{P}$ from $(s, x') \rightarrow (s', x'')$ occurs if the transition is valid in both $\mathcal{M}$ i.e. $P(s, a, s') > 0$, and $\mathcal{D}$, i.e. $x'' = \Delta_{\mathcal{D}_\phi}(x', l(s))$. We can deduce that the construction of $\mathcal{P}$ ensures: 1) the DFA deterministically observes the satisfaction of the given formula based on the observed labels associated with states $l(s_0), l(s_1), \cdots$; 2) for any sequence of states over the MDP $\mathcal{M}$, the probabilistic behavior over the states is preserved, i.e. the probability over the sequence is identical in both MDP $\mathcal{M}$ and $\mathcal{P}$ under any given policy $\rho$. Thus, for any execution of states (i.e., $s_0, s_1, s_2, \cdots$) over the $\mathcal{M}$ under a given policy $\rho$, there exists a corresponding trace over $\mathcal{P}$ (i.e., $(s_0, x_0'), (s_1, x_1'), (s_2, x_2'), \cdots$), where the probability in $\mathcal{M}$ and $\mathcal{P}$, remain unchanged. Hence, the given HyperTWTL formula $\varphi$ is satisfied by a sequence of states in $\mathcal{M}$ if and only if the corresponding label sequence $l(s_0), l(s_1), \cdots$ is accepted in $\mathcal{D}_\phi$. This occurs if and only if an accepting state $F_{\mathcal{D}_\phi}$ is reached in the corresponding sequence of states over $\mathcal{P}$. Therefore, despite the additional states introduced by $\mathcal{D}$ in the construction of $\mathcal{P}$, the transitions in the $\mathcal{P}$, are constrained to transitions that contribute to the satisfaction of the HyperTWTL formula $\varphi$ as encoded in $\mathcal{D}$ and preserve the probabilistic behavior of $\mathcal{M}$. It can then be concluded that the probabilities of satisfaction of HyperTWTL formula $\varphi$ are preserved between MDP $\mathcal{M}$ and product MDP $\mathcal{P}$ in the cross-product construction of $\mathcal{P}$.

**Theorem 3:** Let $\varphi$ be the HyperTWTL formula to be satisfied with a probability of at least $P_{th}$ within the time bound $||\varphi||$. Given an MDP $\mathcal{M}$, Timed MDP $\mathcal{T}$ and some $\varepsilon \in [0, 1)$, we assume each transition over $\mathcal{M}$ exhibits $\varepsilon$−probabilisticity, while any feasible transition over $\mathcal{T}$ can potentially increase the Distance to $F_\mathcal{T}$ by a maximum of one. Given the set of initial states $q_0$ of $\mathcal{T}$ satisfies

$$P_{th} \leq \sum_{j=0}^{\frac{||\varphi|| - F_\mathcal{T}^{dist}(q)}{2}} \frac{||\varphi||}{(||\varphi|| - 1)! j!} \varepsilon^j (1 - \varepsilon)^{||\varphi|| - j}, \forall q \in q_0,$$

then $P(t_1^j, t_2^j, \cdots, t_m^j \models \varphi) \geq P_{th}, \forall j \geq 0$, where $t_1^j, t_2^j, \cdots, t_m^j$ are traces generated from episode $e p^j$, $F_\mathcal{T}^{dist}(q)$ represents the minimum distance from state $q$ to any accepting state in $\mathcal{D}_\mathcal{T}$ under $\varepsilon$-probabilistic transitions, each trace $t_i^j$ represents a path over the MDP following policy $F_\mathcal{T}^{reach}$, and $m$ is the number of trace variables in the given HyperTWTL formula $\varphi$.

*Proof:* According to Algorithm 2, a sequence of $||\varphi||$ actions of each $e p^j$ episode, can be selected by either the $\varepsilon$−greedy approach (Line 10) or softmax function (Line 13). We now define an arbitrary random variable $x_n$ where $x_n = 1$ if an independent action $a_n$ from a sequence of $||\varphi||$ actions with success probability $p = 1 - \varepsilon$. We say an action $a_n$ has success probability if it follows the transitions that contribute to satisfying $\varphi$. We now denote $\mathcal{S} = \sum_{n=1}^{||\varphi||} x_n$ as the total number of successes in $||\varphi||$ steps. Recall, a given formula $\varphi$ is satisfied within the bound $||\varphi||$, if an accepting state is reached from the initial state $q$ within at least $F_\mathcal{T}^{reach}(q)$ successful actions. Assuming $\mathcal{S} \geq F_\mathcal{T}^{dist}(q)$, then it implies that the probability of satisfying $\varphi$ is equivalent to at most $||\varphi|| - F_\mathcal{T}^{dist}(q)$. According to Algorithm 1, the number of failures is bounded below $\frac{||\varphi|| - F_\mathcal{T}^{dist}(q)}{2}$. Given the success of each action is an independent Bernoulli trial with success probability $1 - \varepsilon$, the total number of successful actions $\mathcal{S}$ follows a binomial distribution $B(||\varphi||, 1 - \varepsilon)$. The probability of having at most $\frac{||\varphi|| - F_\mathcal{T}^{dist}(q)}{2}$ failures can be given as

$$P(\mathcal{S} \geq |\varphi| - \frac{||\varphi|| - F_\mathcal{T}^{dist}(q)}{2}) =$$
$$\sum_{j=||\varphi||-|\varphi|-\frac{||\varphi||-F_\mathcal{T}^{dist}(q)}{2}}^{|\varphi|} \binom{||\varphi||}{j} (1 - \varepsilon)^j \varepsilon^{||\varphi||-j}$$

Given $P_{th} \leq \sum_{j=0}^{\frac{||\varphi|| - F_f^{dist}(q)}{2}} \binom{||\varphi||}{j} \varepsilon^j (1 - \varepsilon)^{||\varphi|| - j}, \forall q \in q_0$, we can conclude that the probability of having a number of successes that satisfy the HyperTWTL formula $\varphi$ within the given time frame will be at least $P_{th}$ demonstrating that Algorithm 1 guarantees the satisfaction of the given formula $\varphi$ with a probability of at least $P_{th}$.

**Temporary page!**

LaTeX was unable to guess the total number of pages correctly. As there was some unprocessed data that should have been added to the final page this extra page has been added to receive it.

If you rerun the document (without altering it) this surplus page will go away, because LaTeX now knows how many pages to expect for this document.